# Learning symbol relation tree for online mathematical expression recognition

**Thanh-Nghia Truong**
Dept. of Comp. and Info. Sciences
Tokyo University of Agriculture and Technology (TUAT)
Tokyo, Japan 184-8588
thanhnghiadk@gmail.com

**Hung Tuan Nguyen**
Institute of Global Innovation Research
TUAT
Tokyo, Japan 184-8588
ntuanhung@gmail.com

**Cuong Tuan Nguyen**
Dept. of Comp. and Info. Sciences
TUAT
Tokyo, Japan 184-8588
ntcuong2103@gmail.com

**Masaki Nakagawa**
Dept. of Comp. and Info. Sciences
TUAT
Tokyo, Japan 184-8588
nakagawa@cc.tuat.ac.jp

## Abstract

This paper proposes a method for recognizing online handwritten mathematical expressions (OnHME) by building a symbol relation tree (SRT) directly from a sequence of strokes. A bidirectional recurrent neural network learns from multiple derived paths of SRT to predict both symbols and spatial relations between symbols using global context. The recognition system has two parts: a temporal classifier and a tree connector. The temporal classifier produces an SRT by recognizing an OnHME pattern. The tree connector splits the SRT into several sub-SRTs. The final SRT is formed by looking up the best combination among those sub-SRTs. Besides, we adopt a tree sorting method to deal with various stroke orders. Recognition experiments indicate that the proposed OnHME recognition system is competitive to other methods. The recognition system achieves 44.12% and 41.76% expression recognition rates on the Competition on Recognition of Online Handwritten Mathematical Expressions (CROHME) 2014 and 2016 testing sets.

## 1 Introduction

Mathematical expressions play an essential role in scientific documents since they are indispensable for describing problems, theories, and solutions in math, physics, and many other fields. Due to the rapid emergence of pen-based or touch-based input devices such as digital pens, tablets, and smartphones, people have begun to use the handwriting interfaces as an input method. Although the input method is natural and convenient, it is useful if handwritten mathematical expressions (HMEs) are correctly recognized. There are two approaches to recognize handwriting based on the type of input patterns. The first approach uses the real-time sequences of pen-tip, or finger-top coordinates, collected from modern electronic devices, termed as online input. The other approach processes handwritten images captured from a scanner or a camera termed as offline input.

Generally, offline input is applied for images scanned from handwritten sheets, while online input requires pen-based or touch-based devices. Offline recognition has the problem of segmentation, but it is free from various stroke orders or duplicated strokes. Online input allows touching strokes to be separated from time sequence information, but it is troubled by writing order variation and stroke duplication.

The problem of HME recognition has been studied for decades. Several grammar-based, tree-based, graph-based, or Deep Neural Network (DNN) based approaches have been applied to settle HME recognition [1]–[4]. Generally, three subtasks are involved in both online and offline HME recognition [5], [6]: (1) symbol segmentation to group strokes belonging to the same symbol; (2) symbol recognition to classify the segmented symbol; (3) structural analysis to identify spatial relations between symbols to produce a mathematical interpretation with the help of grammars.

The traditional approach is mainly grammar-driven methods. These methods consist of a set of generated symbol hypotheses and a structural analysis algorithm for selecting the best hypotheses. Nonetheless, they achieve low recognition rates [5], [6] because of the two main reasons. First, they depend on predefined rules and hand-crafted features, which are less robust for various handwriting styles. Second, their isolated subtasks perform poorly without sharing global context. The tree-based and graph-based approaches show their advantages in representing the structures of HMEs. These approaches use a tree [3] or a graph [7] to represent the 2D structure of an HME. However, the approaches are still limited with low HME recognition rates because they also depend on the performance of their isolated subtasks.

The DNN based approach has recently been successfully used to parse the structures directly from HME training samples. It deals with HME recognition as an input-to-sequence problem [4] where input can be either an HME image (OffHME) or an online HME (OnHME) pattern. This approach is flexible and archives good performance because it uses a shared context to learn HME recognition. However, the approach requires a large number of training data to improve the generalization of the DNN-based model. Moreover, DNNs have difficulty extracting the 2D structure of HMEs using the 1D structure of LaTeX sequences [3]. Furthermore, the approach has weak grammar constraints, so that it might generate wrong candidates.

In this work, we focus on OnHME recognition from a new perspective. It is treated as a problem of deriving a symbol relation tree (SRT) from an input OnHME with a temporal classifier. The temporal classifier recognizes an input to produce a sequence of symbols and spatial relations between the recognized symbols. We first use the temporal classifier to generate an SRT and then reconstruct the SRT to get the final SRT representing the input OnHME. The SRT reconstruction method includes a tree-based symbol level sorting, which solves the problem of writing order variation. With this approach, we make the model learn from the 2D structure of the OnHME patterns.

In summary, we make the following contributions:

1. We propose a temporal classifier for OnHME patterns to generate both symbols and relations directly. It helps the model encode a better context for recognizing symbols and predicting the relations between them.

2. We propose an OnHME recognition method based on an SRT that is the output of the temporal classifier.

The rest of the paper is organized as follows. Section II provides an overview of the state-of-the-art. Section III declares our method proposed in this paper. Section IV describes our dataset and experiments. Section V highlights the results of the experiments. Section VI draws the conclusion and future works.

## 2 Related works

In this section, we briefly summarize some methods closely related to our method. These methods basically follow two main steps for symbol recognition and structural analysis.

### 2.1 Structural approach for OnHME recognition

The problem of OnHME recognition, as well as OffHME recognition, has been studied for decades [8] from the early top-down methods [9], bottom-up methods [10], and their variations [5], [6]. The most common approach to represent an OnHME is to use predefined grammars. Yamamoto et al. used Probabilistic Context-Free Grammars [11], MacLean and Labahn developed a method using relational grammars and fuzzy sets [7]. Grammars show their strength in representing the semantic structure of HME [5], [6]. However, the grammars need to be carefully designed to avoid the lack of generality and extensibility. Moreover, the results of symbol segmentation, symbol recognition, and spatial relation classification also affect the result of the grammar-based approach.

Another approach by T. Zhang et al. used Stroke Label Graph (SLG) to represent an OnHME in which nodes represent strokes whereas labels on the edges encode either segmentation or layout information [7]. They used a tree-based model and a sequential model to learn the structure of OnHMEs [3]. The recognition result is obtained by combining the results from different paths to rebuild a math expression. The proposed system did not need grammars for OnHME recognition. However, the result was still limited.

### 2.2 Learning spatial relations

The structural analysis task aims to learn syntactic models that represent spatial relations between the mathematical symbols. Many approaches used a tree-based model in their HME recognition systems [12]–[14] to represent 2D spatial relations of an HME. However, they depend on symbol recognition performance and need to process a large number of hypotheses for spatial relation. Moreover, the approach lacks the global context to classify the spatial relations since they are independently solved with other HME recognition tasks.

The spatial relations could also be learned directly from sequential features [7] or tree-based features [3]. A sequence alignment method named local CTC was proposed in [3] to constraint the spatial relations output to the corresponding time step (i.e., the time steps between two consecutive symbols. This approach benefits from the global context of the sequence or the tree structure. However, their performance is low since learning spatial relations at the stroke level is difficult for the recognizer.

### 2.3 Global context for improving HME recognition

The task of symbol recognition aims at segmenting all symbols in an OnHME and then recognizing the symbols. Generally, many works combined online features extracted from pen traces and offline features from the rendered image of an OnHME [15], [16]. Deep learning-based methods are frequently used to encode DNN-based features for symbol recognition [17].

Graves showed a crucial drawback of isolated classification methods compared with sequence classification ones [18]. The isolated classification methods use only the context information from the current isolated symbol to classify the label. Many other works [6], [15] also demonstrated that the isolated symbol recognition faces difficulty in discriminating symbols of similar shapes such as (X, x, \times), (O, o, 0), without using context.

On the other hand, deep learning-based methods are frequently used to solve HME recognition tasks in recent years. Zhelezniakov et al. used a Bidirectional Long Short-Term Memory (BLSTM) to segment and classify symbols in an HME [19]. For a sequence of symbols in an HME, the BLSTM encodes the global information of the context using the preceding states and succeeding states of a sequential input, and Connectionist Temporal Classification (CTC) optimizes the alignment and classification. Nguyen et al. improved the symbol recognition in an HME by using a deep BLSTM-CTC model to encode global context information [17]. They encoded bidirectional context for symbol classification to solve the problem of recognizing ambiguous symbols. However, the global context is not shared with other HME recognition tasks, such as relation classification.

## 3 Our approach

In this section, we propose a symbol-relation temporal classifier and how we prepare data for training the temporal classifier. Then, we propose an OnHME recognition system using the temporal classifier to generate and then reconstruct an SRT to represent an input OnHME pattern.

### 3.1 Sequential model for segmentation recognition and spatial relation

We propose a sequential model for learning symbol segmentation, symbol recognition, and spatial relations from derived paths of an SRT.

#### 3.1.1. Symbol relation tree (SRT) and derived paths

Each OnHME pattern is represented at multiple levels: expression, symbol, and stroke. Generally, expression-level representation such as LaTeX form is well known. On the other hand, OnHME can be represented at the symbol-level as an SRT or the stroke-level as an SLG. In this work, we focus on the symbol-level representation as an SRT.

When only a syntactic representation of an HME is required, its LaTeX form, a 1D sequence representation, is adequate. However, it is hard to represent the 2D structure of the HME. An operator tree based on symbol operators is an adequate tool to describe its 2D structure without specifying its layout. If the layout is concerned, however, an SRT is more appropriate. An SRT represents the placement of symbols on baselines (writing lines) and the spatial arrangement of the baselines [3].

In the CROHME dataset, there are 101 classes of symbols, including digits, alphabets, operators, and so on. Six spatial relations are also defined in the dataset: Right, Above, Below, Inside (for square root), Superscript, and Subscript. In our experiments, we define a new relation, NoRel, to represent no relation between two unrelated symbols.

Fig. 1 shows an OnHME with its SRT and derived paths. In an SRT, each node represents a symbol, while a label on each edge indicates the relation between symbols. For example, Fig. 1(a) shows a simple HME $\int d^2x$ written in five strokes, where each stroke from [1] to [3] represents a symbol where the stroke [4] and [5] represent the symbol '$x$'. Fig. 1(b) shows the SRT of the HME, the first symbol '$\int$' is the root of the tree; the symbol '$d$' is Right of '$\int$', '$x$' is Right of '$d$' and '2' is Sup of '$d$'. The SRT of the whole structure of the HME could be represented by derived paths of consecutive symbols and spatial relations from the SRT. Fig. 1(c) shows two derived paths of the HME $\int d^2x$. By enumerating all derived paths of an HME, we can reconstruct the SRT from them. Moreover, each derived path is a 1D sequence.

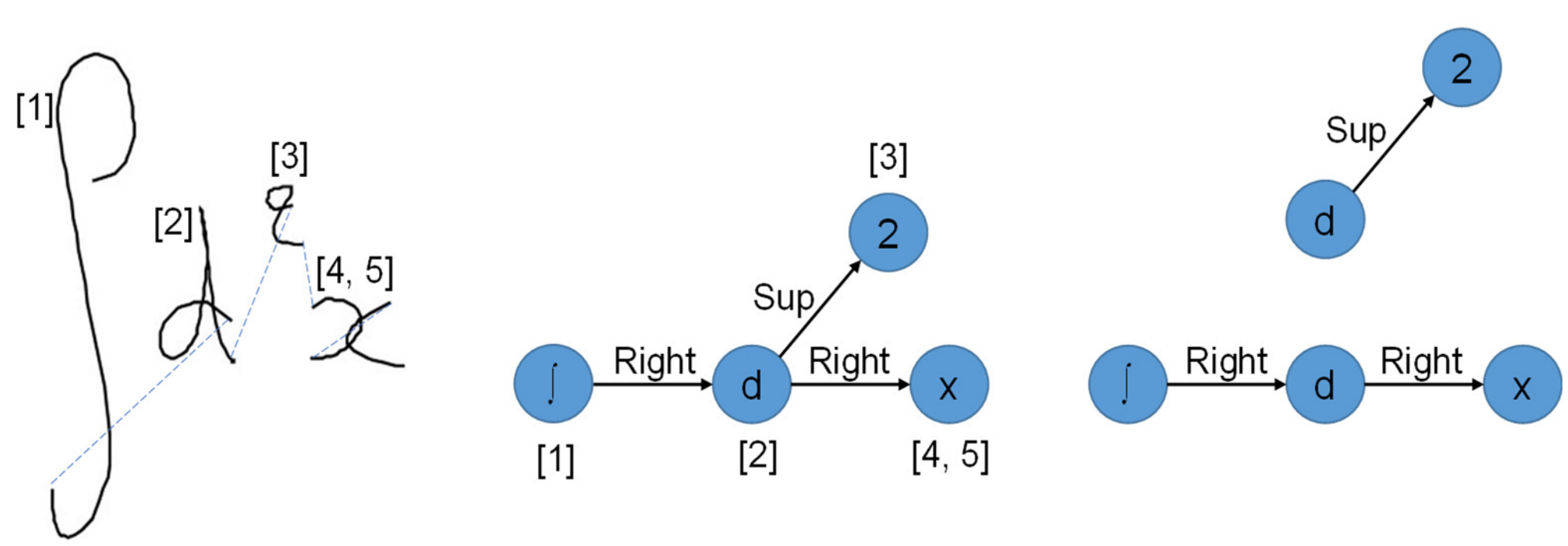

(a) OnHME (b) Symbol relation tree (c) Derived paths

**Fig. 1.** Symbol relation tree and derived paths.

### 3.1.2 Symbol-relation temporal classifier

From this idea, we propose a sequential model for recognizing both symbols and spatial relations in an OnHME. The model receives an input of a feature sequence, consisting of both written strokes and off-strokes (pen movements between strokes), and generates an output sequence of symbols and spatial relations between two consecutive symbols. Assuming that the sequential order of symbols and input strokes is consistent (i.e., there are no delayed strokes), symbols are separated by the off-strokes between them, and they are bound to the corresponding strokes and off-strokes.

We apply a deep BLSTM for symbol classification to encode the global context, as shown in Fig. 2. It combines multiple BLSTM layers that process the input in both forward and backward directions. The forward and backward context by the two LSTM layers is combined and fed to the next BLSTM layer in the networks. The deep BLSTM stacks multiple levels of BLSTM to learn high-level features. Based on the output of the final BLSTM layer, we use a CTC layer to generate symbol segmentation and relation classification results.

From the sequential model, segmentation is performed by finding the off-strokes with a high probability of relations. For an OnHME, let $S$ is a sequence of $n$ strokes of the HME, denoted as $S = (s_0, \dots, s_{n-1})$ and $O$ is a sequence of $(n-1)$ off-strokes as $O = (o_1, \dots, o_{n-1})$ where $o_i$ is an off-stroke between two strokes $s_{i-1}$ and $s_i$. The $i^{th}$ off-stroke $o_i$ is a relation or '**blank**' if it is between two strokes inside a symbol. The relation having the highest probability at the $i^{th}$ off-stroke is considered as the relation between the $(i-1)^{th}$ and $i^{th}$ symbols, as shown in the following Eq. (1):

$$Rel(o_i) = \begin{cases} argmax\left(P_{rel}(o_i|\varphi_{HME})\right) & \text{if } max\left(P_{rel}(o_i|\varphi_{HME})\right) \geq P_{\text{blank}} \\ \textbf{blank} & \text{if } max\left(P_{rel}(o_i|\varphi_{HME})\right) < P_{\text{blank}} \end{cases} \quad (1)$$

where:

- $Rel(o_i)$ is the predicted relation of the $i^{th}$ off-stroke.
- $P_{rel}(o_i|\varphi_{HME})$ is the probability of the relation "$rel$" at $o_i$ given the BLSTM context $\varphi_{HME}$, by parsing $S$ with the symbol-relation temporal classifier.
- $P_{\text{blank}}$ is the probability of $o_i$ being '**blank**' produced by the classifier.

Symbol recognition is performed by taking the maximum probability of symbols between two relation outputs. The symbol recognition for a list of t consecutive strokes $(s_i, \dots, s_{i+t})$ is computed as in Eq. (2):

$$Symbol(s_{\overline{i:i+1}}) = argmax\left(P_{symbol}(o_{\overline{i,i+t}}|\varphi_{HME})\right) \quad (2)$$

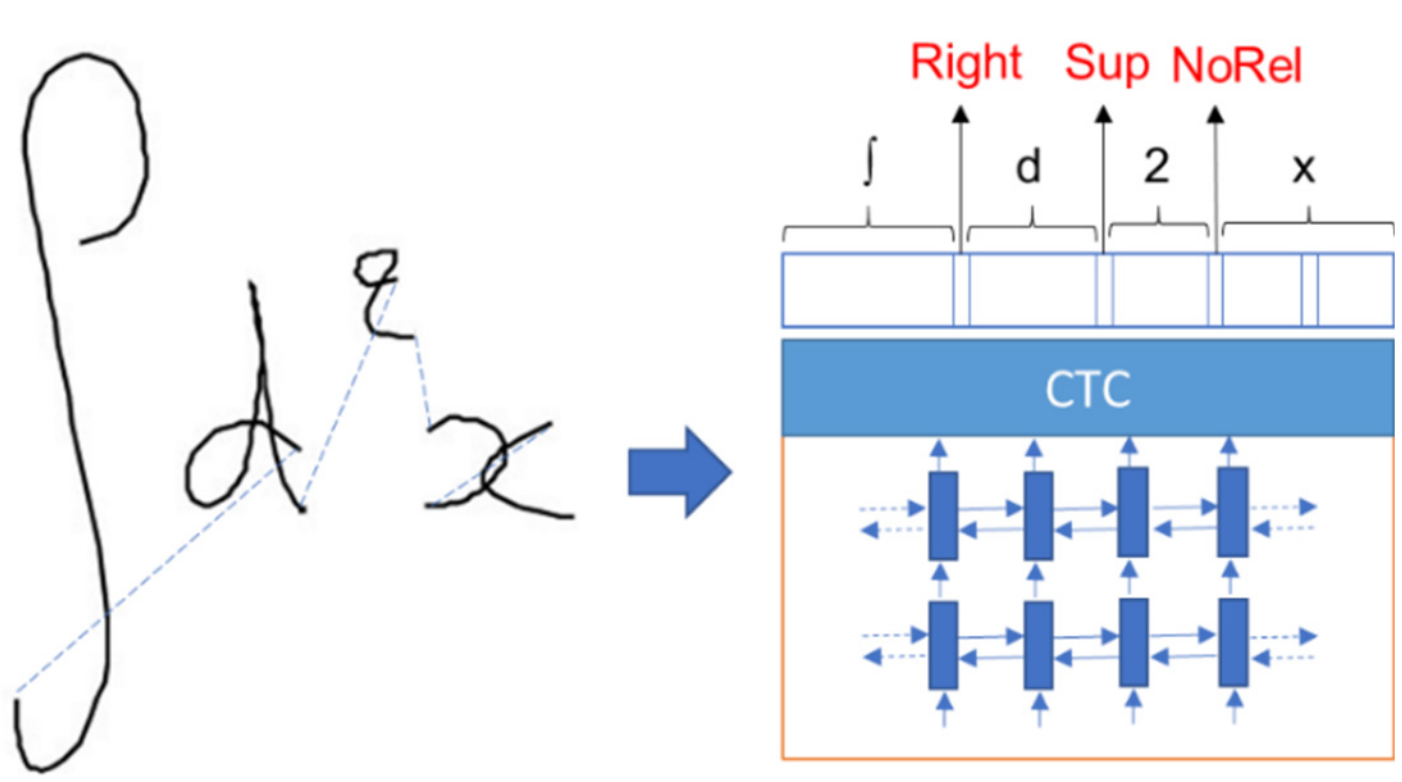

**Fig. 2.** BLSTM for symbol segmentation, recognition, and relation classification.

### 3.1.3 Constraint for precise location output.

Local CTC constrains the output of symbols and relations to the specific strokes and off-strokes [3]. However, the model does not constraint the output of relations to the exact locations, which affects relation classification since the model would be confused on many possible output locations. We use a single feature point representing an off-stroke. We make the model learn precise relations by the constraint loss, which is the combination of the cross-entropy loss function $loss_{CE}$ and the CTC loss function $loss_{CTC}$. The former $loss_{CE}$ is computed as shown in Eq. (3):

$$loss_{CE} = -\sum_{i=0}^{n-1} log(1 - \sum P_{rel}(s_i|\varphi_{HME})) \quad (3)$$

and the combined loss function to train the model is computed as in Eq. (4):

$$loss = loss_{CTC} + loss_{CE} \quad (4)$$

## 3.2 Training path extraction

As we mentioned above, there are multiple derived paths in the SRT for an OnHME. From the SRT, we extract all of them, with each representing a path of strokes and off-strokes as well as their corresponding labels to prepare data for training the symbol-relation temporal classifier.

We propose the following three path extraction rules (***PE-rules***):

- ***PE-rule*** 1: trace all paths from the root to the leaves of an SRT.
- ***PE-rule*** 2: trace the path by writing order. When there is no relation between two consecutive nodes, NoRel is added.
- ***PE-rule*** 3: extract random paths. The rule simulates various writing orders by randomly shuffling the order of sub-trees connected to the root node.

***PE-rule*** 1 extracts all the spatial relations along a path except NoRel. ***PE-rule*** 2 and ***PE-rule*** 3 assign NoRel when tracing a derived path containing two consecutive symbols without a spatial relation between them. Fig. 3 shows an example of applying ***PE-rule*** 1 and ***PE-rule*** 2 to produce the derived paths from the root and the derived path by writing order. Fig. 4 shows an example of applying ***PE-rule*** 3 to generate one of the random paths. Note that ***PE-rule*** 3 generates more training patterns by shuffling all sub-SRTs, which is hard for other LaTeX based methods.

## 3.3 Tree reconstruction

For a list of strokes in an OnHME pattern, the temporal classifier generates an SRT in 1D as a sequence of several derived paths. Each derived path represents the symbol segmentation, and relation classification results sequentially based on the order of input strokes. In the SRT, two consecutive symbols have a suitable relation between them. Otherwise, no relation is displayed. Therefore, we cut the tree into multiple sub-SRTs at the point that the relation is NoRel and then connect these discrete sub-SRTs to build up a final SRT that represents the whole OnHME.

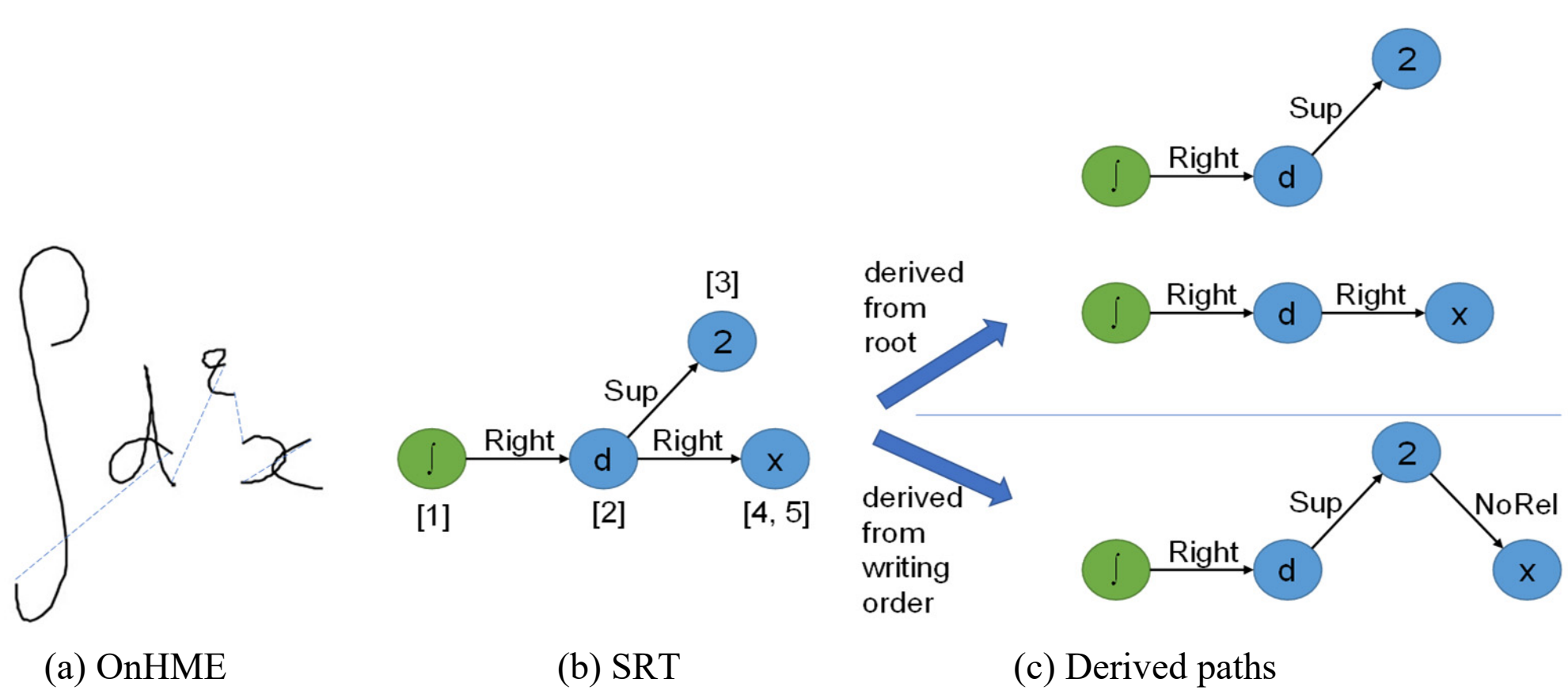

**Fig. 3.** Paths derived from the SRT: derived from the root, or from writing order.

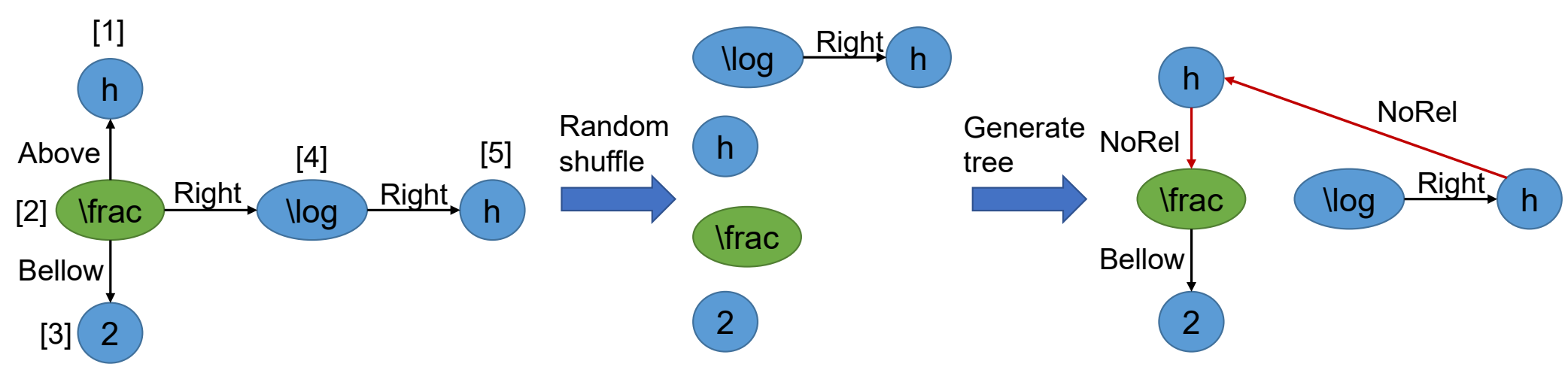

(a) SRT with root is "¥frac" (b) Shuffled sub-SRTs (c) Generated random path

**Fig. 4.** Random path generation.

The temporal classifier recognizes the whole input sequence into the sequence of symbols and spatial relations based on the order of them in the input sequence. First, sub-SRTs are obtained by splitting the temporal classification results of symbols and spatial relations. Then, sub-SRTs are connected into the SRT by additional spatial relations classification by the temporal classifier. Fig. 5 illustrates our recognition method for an HME pattern: $\frac{h}{2}\log h$.

Based on the idea, we propose a tree connector that recognizes an OnHME pattern. Algorithm 1 presents the precise details on how to recognize an input OnHME using the temporal classifier. For an input OnHME pattern, we first get a sequence of strokes $S = (s_0, \ldots, s_{n-1})$ and then pass it to the symbol-relation temporal classifier to recognize the 1D SRT, a sequence of symbols, and relations between them. Next, we cut the 1D-SRT into sub-SRTs and sort them by the information of their bounding boxes. Finally, we connect the discrete sub-SRTs using local and global connections.

To solve the writing order variation problem, we propose a method to sort the sub-SRTs before connecting them. We sort the sub-SRTs using their bounding box coordinates by the following sorting rules (***S-rules***) as shown in Fig. 6:

- ***S-rule*** 1: If a sub-SRT A is completely in the left position of another sub-SRT B, A is sorted before B.
- ***S-rule*** 2: If ***S-rule*** 1 is not matched and a sub-SRT A is completely in the above position of another sub-SRT B, A is sorted before B.
- ***S-rule*** 3: If ***S-rule*** 1 and 2 are not matched, the sub-SRTs are sorted along the x-axis.

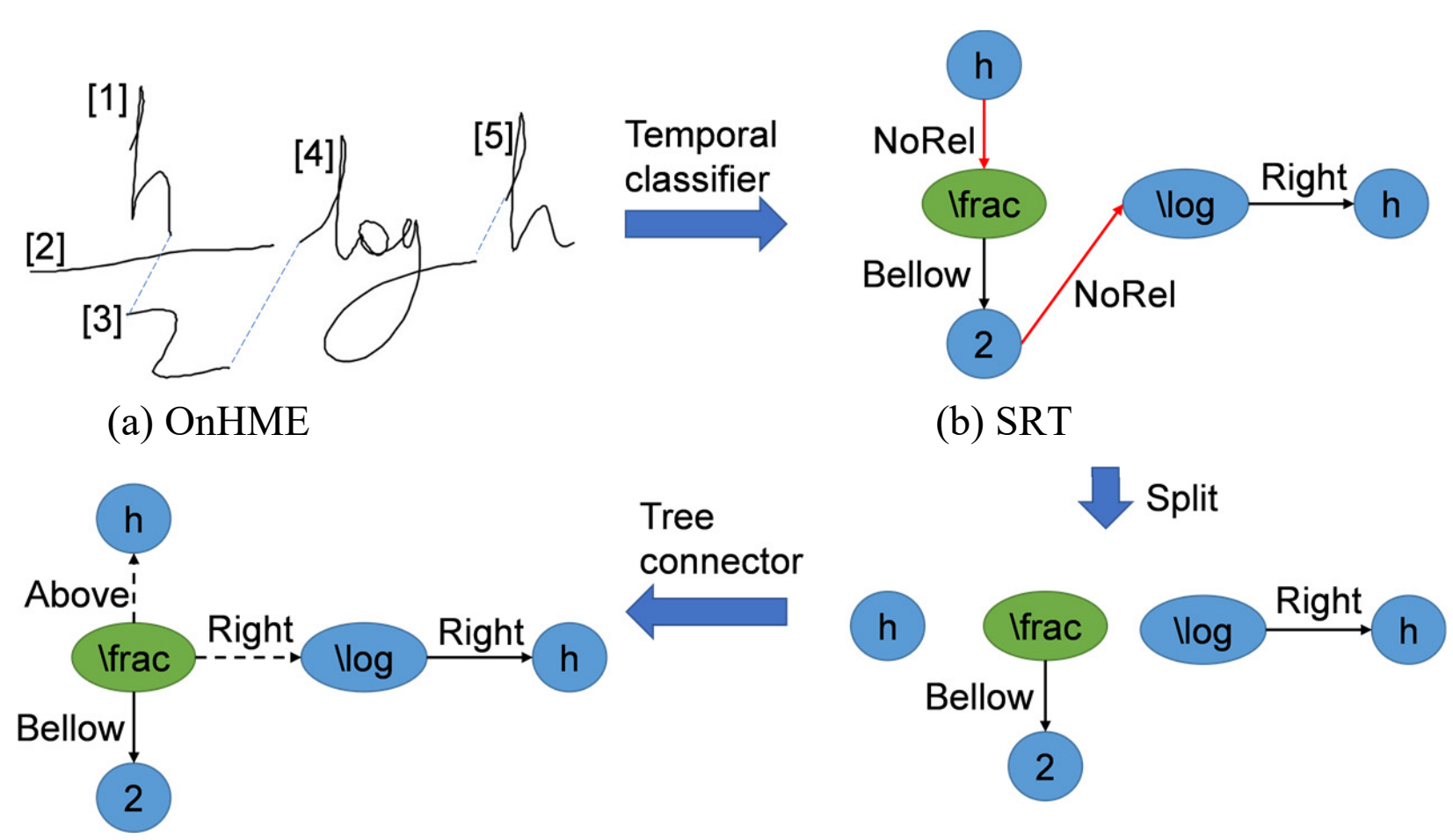

(a) OnHME (b) SRT

(d) Final SRT (c) sub-SRTs

**Fig. 5.** Processes of recognizing an HME pattern: $\frac{h}{2}\log h$.

**Algorithm 1**: Graph cut and connection

**Input:**

A sequence of strokes $S = (s_0, \dots, s_{n-1})$ and a trained symbol-relation temporal classifier (SRTC).

**Output:**

SRT describing the OnHME.

Let L is the SRT output of the temporal classifier: $L = SRTC(s_0, \dots, s_{n-1})$
Let $L_2$ is the list of sub-SRTs when cutting L at the point that there is NoRel between two symbols.
Sort $L_2$ based on its component's position.
**repeat**
  **for** i = 1, …, length ($L_2$) – 1 **do**
    Let $l_i$ is the $i^{th}$ sub-SRT in $L_2$
    Let m is the length of $L_2$
    #local_connect($l_i$, $l_{i+1}$)
    Let $L_3$ is the list of leaf nodes or nodes that do not have Right child in $l_i$
      Let $relation_i = \underset{n_j \, in \, L_3}{\operatorname{argmax}} P_{rel}\left(n_j, l_{i+1} | SRTC(n_j, l_{i+1})\right)$.
    If $relation_i$ is valid, connect $l_i$ with $l_{i+1}$, remove $l_{i+1}$ from $L_2$, break.
    #global_connect($l_i$, $l_{1:i,\ i+1:length(L_2)}$)
    Let $relation_i = \underset{n_j \, in \, L_3, k=[1:i,\ i+1:length(L_2)]}{\operatorname{argmax}} P_{rel}\left(n_j, l_k | SRTC(n_j, l_k)\right)$.
    If $relation_i$ is valid, connect $l_i$ with $l_{i+1}$, remove $l_{i+1}$ from $L_2$, break.
    #can not connect
    flag($l_i$) = True
  **end**
**until** length ($L_2$) = 1 or all flag of $L_2$ is True
**return** $L_2$

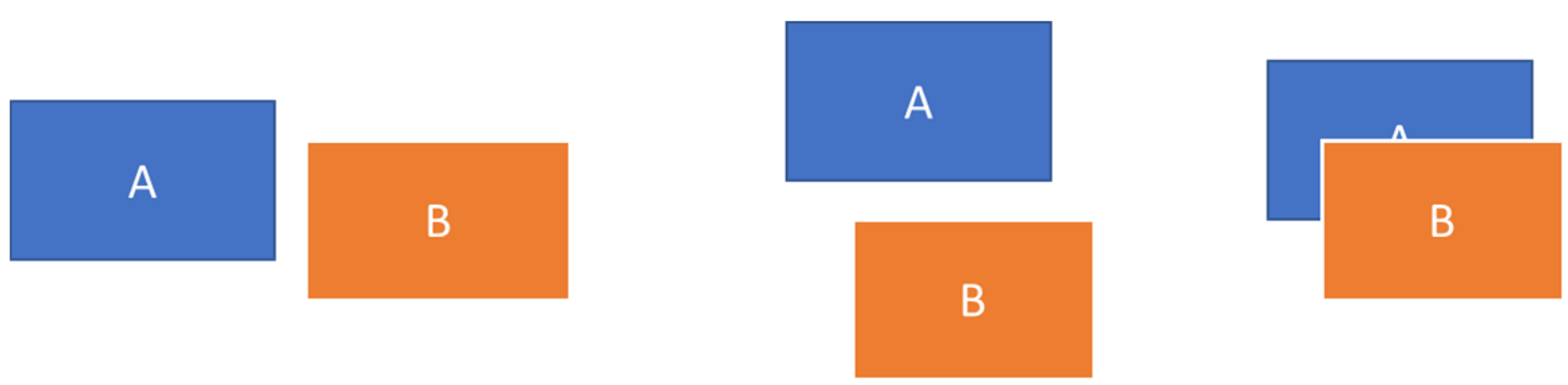

(a) Rule 1　　(b) Rule 2　　(c) Rule 3

**Fig. 6.** Position between two sub-SRTs.

The sorting step solves the problem of writing order variation at the symbol-level. For example, the symbol-relation temporal classifier might make a false symbol recognition or redundant recognition for a symbol with non-consecutive strokes (a part of the symbol written after). There is room to improve the performance by solving the writing order problem at the stroke level.

In order to build the final SRT from the sorted list of sub-SRTs, the tree connector verifies whether two sub-SRTs are appropriate to connect by a local or global connection and then connect them. For each sub-SRT, the local connection denotes that the tree connector finds its adjacent sub-SRTs in the sorted list to connect them. If it does not have any adjacent sub-SRT, the tree connector considers all other sub-SRTs as the global connection. Next, the tree connector looks up all the symbols without the Right node (candidate symbols) in the first sub-SRT. Then, the temporal classifier predicts the relation probabilities between the candidate symbols and the second sub-SRT. Consequently, the highest relation probability indicates the best pair of a candidate symbol in the first sub-SRT and the second sub-SRT.

# 4 Evaluation

## 4.1 Dataset

Data samples and evaluation metrics were introduced in CROHME [5], [6]. Revisions of the datasets and metrics from CROHME 2011 to CROHME 2016 were made for more precise evaluation. We use the CROHME 2014 training set to train the proposed model. For model validation and selection, we use the CROHME 2013 testing set. For evaluation, we use the CROHME 2014 and 2016 testing sets.

In our experiments, we use the CROHME tool LgEval to evaluate our proposed model. We mainly use the expression recognition rate (ExpRate) for the metrics, which shows the ratio of successfully recognized OnHMEs over all OnHMEs. We also use the Precision and Recall rates provided by the LgEval tool [6].

### 4.2 Experiments

We evaluate the effects of the data generation method and the mathematical LM. We compare our OnHME recognition system with the other systems on the CROHME 2014 and 2016 testing sets. We show the effectiveness of the proposed system by comparing it with the benchmark proposed in [3].

#### 4.2.1 Result of HME recognition.

In this section, we evaluate the proposed method on the CROHME 2014 and 2016 testing sets. Table 1 and Table 2 show the evaluation results on the CROHME 2014 testing set at the symbol level and expression level, respectively. Then, Table 3 and Table 4 show those on the CROHME 2016 testing set, respectively. Table 1 and Table 3 consist of recall and precision rates for symbol segmentation (Segments), symbol segmentation and recognition (Segment+Class), and spatial relation classification (Tree relations). A spatial relation prediction between two symbols is correct if the relation is correctly classified and the two symbols are correctly segmented and recognized. Tables 2 and 4 compare the HME recognition rates of the proposed methods with the other methods on the CROHME 2014 and 2016 testing sets, respectively.

**Table 1.** Symbol level evaluation on CROHME 2014 testing set.

| System | Segments (%) | | Segment + Class (%) | | Tree relations (%) | |
|---|---|---|---|---|---|---|
| | Recall | Precision | Recall | Precision | Recall | Precision |
| MyScript [5] | 98.42 | 98.13 | 93.91 | 93.63 | 94.26 | 94.01 |
| Valencia [5] | 93.31 | 90.72 | 86.59 | 84.18 | 84.23 | 81.96 |
| Nantes [5] | 89.43 | 86.13 | 76.53 | 73.71 | 71.77 | 71.65 |
| RIT-CIS [5] | 88.23 | 84.20 | 78.45 | 74.87 | 61.38 | 72.70 |
| RIT-DRPL [5] | 85.52 | 86.09 | 76.64 | 77.15 | 70.78 | 71.51 |
| Tokyo [5] | 83.05 | 85.36 | 69.72 | 71.66 | 66.83 | 74.81 |
| Sao Paulo [5] | 76.63 | 80.28 | 66.97 | 70.16 | 60.31 | 63.74 |
| Tree BLSTM* [3] | 95.52 | 91.31 | 89.55 | 85.60 | 78.08 | 74.64 |
| Our system | 87.69 | 98.11 | 82.62 | 92.44 | 82.28 | 93.02 |

"*" denotes an ensemble of multiple recognition models.

**Table 2.** Expression level evaluation on CROHME 2014 testing set.

| System | Correct (%) | ≤ 1 error | ≤ 2 errors | ≤ 3 errors |
|---|---|---|---|---|
| MyScript [5] | 62.68 | 72.31 | 75.15 | 76.88 |
| Valencia [5] | 37.22 | 44.22 | 47.26 | 50.20 |
| Nantes [5] | 26.06 | 33.87 | 38.54 | 39.96 |
| Tokyo [5] | 25.66 | 33.16 | 35.90 | 37.32 |
| RIT-DRPL [5] | 18.97 | 28.19 | 32.35 | 33.37 |
| RIT-CIS [5] | 18.97 | 26.37 | 30.83 | 32.96 |
| Sao Paulo [5] | 15.01 | 22.31 | 26.57 | 27.69 |
| TAP* [4] | 61.16 | 75.46 | 77.69 | 78.19 |
| Tree BLSTM*[3] | 29.91 | 39.94 | 44.96 | 50.15 |
| Our system | 44.12 | 52.94 | 56.29 | 58.62 |

"*" denotes an ensemble of multiple recognition models.

In Table 1 and Table 2, the top-ranked system MyScript is built on the principle that segmentation, recognition, and interpretation have to be handled concurrently at the same level to produce the best candidate. They used a large number of extra training samples of HME patterns. System Valencia parsed expressions using two-dimensional stochastic context-free grammars. System TAP [4] used an end-to-end network to learn directly LaTeX sequences from OnHME patterns. Tree BLSTM [3] is the benchmark to compare with our system, where they used a tree-BLSTM-based recognition system. Details of works in Table 1 and Table 2 can be found in the CROHME 2014 [5].

For symbol segmentation, recognition, and relation classification, our system archives better precision than the benchmark with nearly equal to the MyScript team. For relation classification, our system achieves 4.21 points higher recall rate with 82.28% compared to 78.07% of the benchmark. However, the recall of our system is not better than the benchmark for symbol segmentation and recognition with around 8 points lower. Our system reaches the level between the second-ranked and the third-ranked systems in CROHME 2014. The global expression recognition rate is 44.12%, ranking third in all systems.

In Table 3 and Table 4, MyScript used a large number of extra training samples of HME patterns, as mentioned above. The team Wiris won the CROHME 2016 competition, but in their work, they trained a language model using a Wikipedia formula corpus consisting of more than 592,000 formulas. Details of other works in Table 3 and Table 4 can be found in the CROHME [6].

For the symbol level evaluation, our system accounts for a competitive result compared with the other participating systems and the benchmark Merge 9, where the precision stands at the level between the first and the second team. However, the recall is still limited as compared with the other teams and the benchmark. There is room for improvement.

TAP [4] achieved the best OnHME recognition rate with 57.02% among the systems without using extra samples for the expression levels. Our system accounts for the ExpRate of 41.76%, a competitive result with the other participant systems in the contest without extra samples.

**Table 3.** Symbol level evaluation on CROHME 2016 testing set.

| System | Segments (%) | | Segment + Class (%) | | Tree relations (%) | |
|---|---|---|---|---|---|---|
| | Recall | Precision | Recall | Precision | Recall | Precision |
| MyScript | 98.89 | 98.95 | 95.47 | 95.53 | 95.11 | 95.11 |
| Wiris | 96.49 | 97.09 | 90.75 | 91.31 | 90.17 | 90.79 |
| Tokyo | 91.62 | 93.25 | 86.05 | 87.58 | 82.11 | 83.64 |
| Sao Paulo | 92.91 | 95.01 | 86.31 | 88.26 | 81.48 | 84.16 |
| Nantes | 94.45 | 89.29 | 87.19 | 82.42 | 73.20 | 68.72 |
| Tree BLSTM* [3] | 95.64 | 91.44 | 89.84 | 85.90 | 77.23 | 74.08 |
| Our system | 88.78 | 97.85 | 83.53 | 92.06 | 82.85 | 92.26 |

"*" denotes an ensemble of multiple recognition models.

**Table 4.** Expression level evaluation on CROHME 2016 testing set.

| System | Correct (%) | ≤ 1 error | ≤ 2 errors | ≤ 3 errors |
|---|---|---|---|---|
| MyScript | 67.65 | 75.59 | 79.86 | |
| Wiris | 49.61 | 60.42 | 64.69 | |
| Tokyo | 43.94 | 50.91 | 53.70 | |
| Sao Paulo | 33.39 | 43.50 | 49.17 | |
| Nantes | 13.34 | 21.02 | 28.33 | |
| TAP* [4] | 57.02 | 72.28 | 75.59 | 76.19 |
| Tree BLSTM* [3] | 27.03 | 35.48 | 42.46 | 27.03 |
| Our system | 41.76 | 49.43 | 52.40 | 54.84 |

"*" denotes an ensemble of multiple recognition models.

### 4.2.2 Error analysis

In this section, we make an in-depth error analysis of our system's recognition results to understand better and explore the directions for improving recognition rate in the future using the CROHME validation tool [6].

Table 5 lists the types of node label errors by our system where the number of errors is larger than or equals 10 on the CROHME 2014 testing set. The first column gives the output node labels by the classifier; the second column provides the ground truth node labels and the number of nodes with each label; the last column records the number of occurrences and the percentages. As can be seen from the table, the most frequent error (x → X, 36) belongs to the type of lowercase-uppercase errors. Moreover, (P → p, 22), (C → c, 16), and (Y → y, 16) also belong to the same type of lowercase-uppercase errors. Another type of common errors is the confusion between the symbols having similar shapes, such as (\times → x, 20), (\ldots → ., 13), (\div → +,12), (1 → COMMA, 12), and so on. Another improvement would be to integrate a language model explicitly to promote frequent symbols.

Table 6 provides the edge label errors on the CROHME 2014 testing set using our OnHME recognition system. The first column represents the output labels; the first row offers the ground truth labels, the number of edges with each label; the other cells in this table provide the corresponding no. of occurrences. '*' represents segmentation edges within a symbol.

Since the Right edges are the most common edges among six spatial relations, the number of confusions to Right edges is the largest among the spatial relations. However, there are few confusions of Above and Below edges to Right edges and vice versa. Additionally, there are only 11 over 377 Inside edges being confused to Right edges. The result shows that our temporal classifier could learn spatial relations effectively with the global context.

**Table 5.** Node label errors of our system on CROHME 2014 testing set.

| Output label | Ground truth label (no. of nodes with this label) | No. of occurrences (percentage) |
|---|---|---|
| X | x (890) | 36 (4.04%) |
| P | p (76) | 22 (28.95%) |
| \times | x(890) | 20 (2.25%) |
| C | c (90) | 16 (17.78%) |
| Y | y (223) | 16 (7.17%) |
| \ldots | . (18) | 13 (72.22%) |
| \div | + (599) | 12 (2.0%) |
| 1 | COMMA (87) | 12 (13.79%) |
| 2 | z (116) | 11 (9.48%) |
| V | v (53) | 11 (20.75%) |
| 9 | q (25) | 10 (40.0%) |
| S | s (25) | 10 (40.0%) |
| \lamda | x (890) | 10 (1.12%) |

**Table 6.** Edge label errors of our system on CROHME 2014 testing set.

| G-truth / Input | * (9044) | Above (592) | Below (627) | Inside (377) | Right (13698) | Sub (1115) | Sup (923) | NoRel (261528) |
|---|---|---|---|---|---|---|---|---|
| * | 486 | 2 | 1 | 2 | 33 | 3 | 1 | 76 |
| Above | 5 | | | | 5 | | | 52 |
| Below | 5 | | | | | 1 | | 17 |
| Inside | | 3 | 3 | | | | | 5 |
| Right | 65 | | | 11 | | 103 | 22 | 593 |
| Sub | 3 | | 1 | 1 | 43 | | 2 | 9 |
| Sup | 3 | 1 | | | 18 | | | 32 |
| NoRel | 1017 | 189 | 193 | 35 | 2021 | 174 | 191 | |

The current system still encounters the problem of missing relations or determine them as NoRel. Two reasons could produce the problem. First, the temporal classifier may fail to detect symbols or misclassified spatial relations into NoRel. The other reason is that the proposed method might skip some sub-SRTs in the SRT reconstruction step in some cases where the system cannot connect two sub-SRTs with local and global connections. Moreover, this is also why the precision rate is high, but the recall rate is low on the CROHME testing sets. Therefore, there is room to improve the tree reconstruction method.

## 5 Conclusion

In this work, we proposed an LSTM-based temporal classifier that used a shared context to learn from OnHME patterns to solve symbol recognition and relation classification. We used the temporal classifier to build a tree-based OnHME recognition system. Our recognition system has achieved the expression recognition rates of 44.12% and 41.76% on the CROHME 2014 and 2016 testing sets, respectively. They are inferior to the best recognition rates but better than the related approach of 29.91% and 27.03% by the previous tree-based OnHME recognition system [3]. The proposed system learns better relation classification with higher recognition rates, and there is no massive drop in the Recall rate of relation classification. We plan to improve the symbol-relation temporal classifier by sorting the strokes to avoid the stroke order variation problem in our future work.

**Acknowledgment**
This work is being supported by the Grant-in-Aid for Scientific Research (A)-18H03597 and that for Early Career Scientist-18K18068.